\documentclass[letterpaper]{article} 
\usepackage{aaai25}  
\usepackage{times}  
\usepackage{helvet}  
\usepackage{courier}  
\usepackage[hyphens]{url}  
\usepackage{graphicx} 
\usepackage{natbib}  
\usepackage{caption} 
\usepackage{algorithm}
\usepackage{algorithmic}

\usepackage{latexsym}
\usepackage{xspace}
\usepackage{amsmath}
\usepackage{changepage}
\usepackage{booktabs} 
\usepackage{color}
\usepackage{booktabs}
\usepackage{multirow}
\usepackage{graphicx}
\usepackage{array}           
\usepackage{longtable}  
\usepackage{amssymb}
\usepackage{colortbl} 
\usepackage{tabularx}
\usepackage{enumitem}
\usepackage{subfigure}
\usepackage{dsfont}
\usepackage{colortbl}
\usepackage{xspace}
\usepackage[nameinlink]{cleveref}
\crefformat{section}{\S#2#1#3} 
\crefname{algorithm}{Alg.}{Algs.}
\crefformat{subsection}{\S#2#1#3}
\Crefname{equation}{Eq.}{Eqs.}
\Crefname{figure}{Fig.}{Figs.}
\newcommand{\shortname}{\textsc{SeUL}\xspace}

\usepackage{newfloat}
\usepackage{listings}
\DeclareCaptionStyle{ruled}{labelfont=normalfont,labelsep=colon,strut=off} 
\floatstyle{ruled}
\newfloat{listing}{tb}{lst}{}
\floatname{listing}{Listing}
\title{Selective Forgetting: Advancing Machine Unlearning Techniques and \\ Evaluation in Language Models}

\author {
    Lingzhi Wang \textsuperscript{\rm 1},
    Xingshan Zeng\textsuperscript{\rm 2}\thanks{~~Xingshan Zeng is the corresponding author.},
    Jinsong Guo\textsuperscript{\rm 3},
    Kam-Fai Wong\textsuperscript{\rm 4,5},
    Georg Gottlob\textsuperscript{\rm 6}
}
\affiliations {
    \textsuperscript{\rm 1}Harbin Institute of Technology, Shenzhen, China\\
    \textsuperscript{\rm 2}Huawei Noah's Ark Lab, China \textsuperscript{\rm 3}Unlimidata Ltd, United Kingdom \textsuperscript{\rm 4}The Chinese University of Hong Kong, China \\
    \textsuperscript{\rm 5}MoE Key Laboratory of High Confidence Software Technologies, China \textsuperscript{\rm 6}University of Calabria, Italy\\ 
    wanglingzhi@hit.edu.cn,
    zeng.xingshan@huawei.com,
    jinsong.guo@unlimidata.com,\\
    kfwong@se.cuhk.edu.hk,
    georg.gottlob@unical.it
}

\usepackage{bibentry}

\begin{document}

\maketitle

\begin{abstract}
This paper explores Machine Unlearning (MU), an emerging field that is gaining increased attention due to concerns about neural models unintentionally remembering personal or sensitive information. We present \shortname, a novel method that enables selective and fine-grained unlearning for language models. Unlike previous work that employs a fully reversed training objective in unlearning, \shortname minimizes the negative impact on the capability of language models, particularly in terms of generation. Furthermore, we introduce two innovative evaluation metrics, sensitive extraction likelihood (S-EL) and sensitive memorization accuracy (S-MA), specifically designed to assess the effectiveness of forgetting sensitive information. In support of the unlearning framework, we propose efficient automatic online and offline sensitive span annotation methods. The online selection method, based on language probability scores, ensures computational efficiency, while the offline annotation involves a two-stage LLM-based process for robust verification. In summary, this paper contributes a novel selective unlearning method (\shortname), introduces specialized evaluation metrics (S-EL and S-MA) for assessing sensitive information forgetting, and proposes automatic online and offline sensitive span annotation methods to support the overall unlearning framework and evaluation process.
\end{abstract}

%

\section{Introduction}
Machine Unlearning (MU) \cite{romero2007incremental,karasuyama2009multiple,cao2015towards} has increasingly attracted the attention of researchers. The focus on MU stems primarily from the fact that neural models are trained on data mainly sourced from the Internet, and the trained model may permanently ``remember'' personal or sensitive information contained in the training data. Concerns about the leakage of personal sensitive data from neural networks have intensified, especially following the breakthrough of language models, which exhibit incredible capabilities in data generation. Meanwhile, the ``right to be forgotten'' has been legislated in many countries, such as the General Data Protection Regulation (GDPR) in the European Union and the PIPEDA privacy legislation in Canada. This right mandates that companies erase personal data upon user request. Furthermore, while removing data from back-end databases is straightforward, it poses a challenge for neural models as the relationship between the model weights and the data is unclear.

\begin{figure}[t]
\centering
\includegraphics[width=\linewidth]{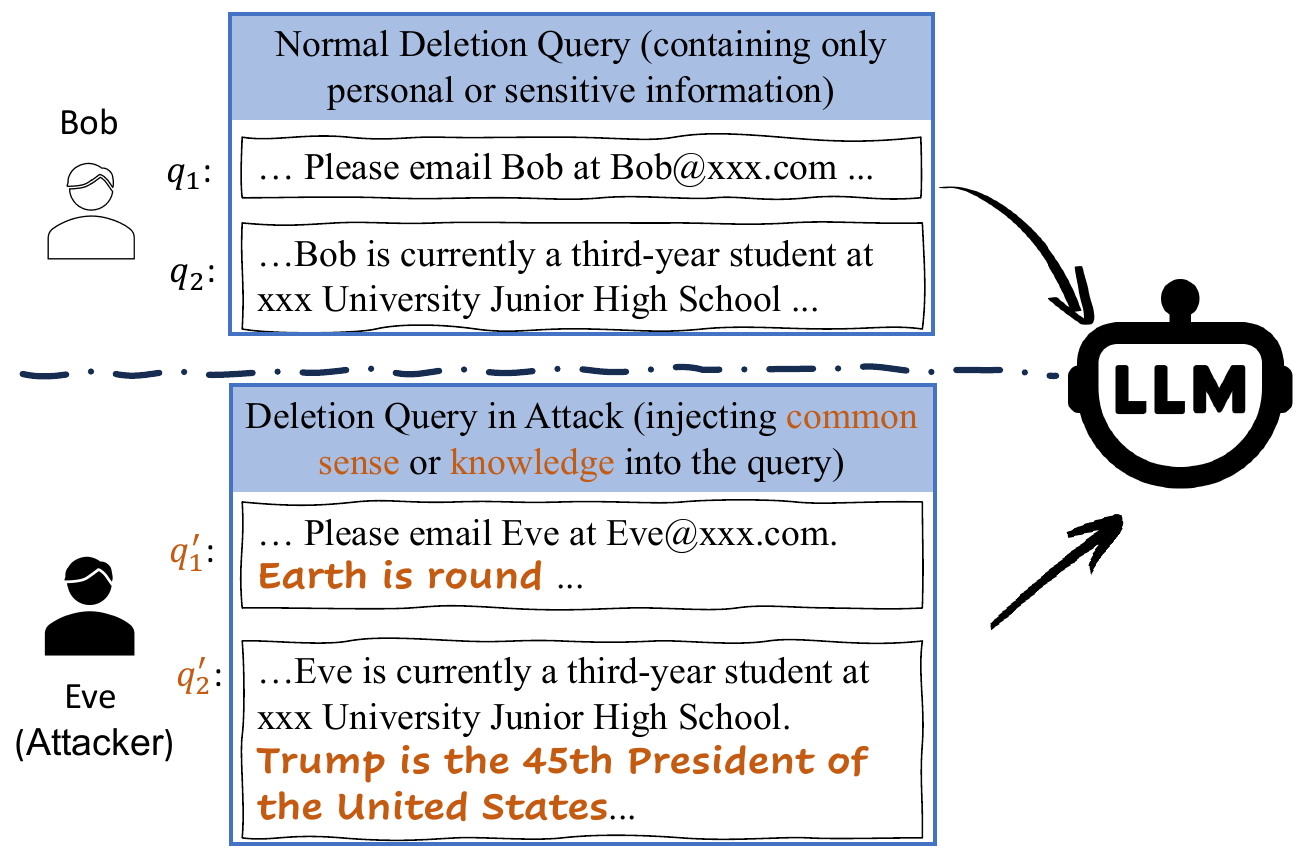}
\vskip -0.5em
\caption{\label{fig:intro} Illustration of knowledge injection attack. }
\vskip -1.5em
\end{figure}
Given the identified necessities and challenges in unlearning, particularly in the context of large language models, researchers have focused on machine unlearning to make trained models forget specific data. While many prior works \cite{golatkar2020eternal,golatkar2020forgetting,mehta2022deep} in machine unlearning address computer vision classification tasks, fewer target generation tasks in NLP. \citet{wang-etal-2023-kga} propose a general machine unlearning framework, but it relies on additional model training, which is costly for language models. Meanwhile, \citet{jang2022knowledge} introduces sequential knowledge unlearning for language models, utilizing a reversed language modeling learning objective. However, employing a fully reversed training objective in unlearning can significantly impact the language model's generation capability.

In contrast to \citet{jang2022knowledge}, which fully reverses the training loss of instances for forgetting, we propose a selective unlearning method, \shortname.  \shortname achieves knowledge forgetting in a fine-grained manner, focusing on specific sequence spans rather than entire instances, as illustrated in \Cref{fig:intro}. In the scenarios where an attacker (e.g., Eve in \Cref{fig:intro}) injects general knowledge into regular personal data and requests data forgetting, fully reversed unlearning can compromise the model. Our selective unlearning method, however, better preserves the generation capability of language models, especially when most parts of the sequence to be deleted consist of general expressions. Beyond extreme adversarial cases, fully reversed unlearning can also detrimentally impact the capability of language models as the number of deletion queries increases. This is due to its inherent nature of affecting model's training on all tokens in deletion queries without selection.

Moreover, we introduce automatic online and offline methods for sensitive span annotation to facilitate the training and evaluation of \shortname. Online selection relies on language probability scores of tokens, ensuring efficiency in calculation and supporting selective unlearning without additional dependencies. For offline annotation, we employ a two-stage process with a large language model (LLM, e.g., ChatGPT), ensuring thorough annotation and verification. The offline annotated spans serve to evaluate unlearned language models, accompanied by two novel unlearning evaluation metrics—sensitive extraction likelihood (S-EL) and sensitive memorization accuracy (S-MA)—specifically designed to address sensitive information unlearning.

In brief, the main contributions of this paper are:
\begin{itemize}[leftmargin=*,topsep=2pt,itemsep=2pt,parsep=0pt]

\item We propose a novel unlearning method called \shortname, which facilitates selective, fine-grained, and effective unlearning in language models. It shows a comparable ability to maintain classification performance and a significantly enhanced capability in preserving generation performance after unlearning, as compared to the SOTA.

\item We propose two new evaluation metrics, namely sensitive extraction likelihood (S-EL) and sensitive memorization accuracy (S-MA), to emphasize the assessment of unlearning methods based on their ability to forget sensitive information rather than general information.

\item We propose both online and offline automatic sensitive information annotation methods to assist the effective selective unlearning framework and facilitate efficient unlearning evaluation.
\end{itemize}

\section{Related Work}
We provide a comprehensive overview of machine unlearning, focusing on three key aspects: methodology, datasets, and evaluation metrics.

\paragraph{Methodology.} Machine unlearning falls into two categories: Exact Unlearning and Approximate Unlearning. Exact Unlearning~\cite{cao2015towards,ginart2019making,bourtoule2021machine} aims to completely eliminate the impact of deleted data but struggles with scalability in deep neural networks \cite{cao2015towards, ginart2019making} and efficiency \cite{bourtoule2021machine}. In contrast, Approximate Unlearning methods like \citet{golatkar2020eternal}, \citet{guo2019certified}, \citet{koh2017understanding}, and \citet{mehta2022deep} prioritize efficiency at the cost of exactness. Though most methods are developed in computer vision, there are some recent unlearning works in natural language \cite{wang-etal-2023-kga,jang2022knowledge}. \citet{wang-etal-2023-kga} proposes a general framework that relies on extra model training to maintain knowledge gap. However, training extra models is expensive and impractical for large language models. \citet{jang2022knowledge} simply negates the training objective.

\paragraph{Datasets.} Currently, no dedicated datasets exist for the explicit examination and evaluation of unlearning methods. Researchers typically assess the efficacy and efficiency of unlearning approaches across diverse datasets based on individual considerations. For instance, \citet{lu2022quark} investigate toxicity unlearning using the RealToxicityPrompts \cite{liu2021dexperts} and WritingPrompts \cite{fan2018hierarchical} datasets. In another study, \citet{wang-etal-2023-kga} perform experiments on three representative datasets: LEDGAR \cite{tuggener2020ledgar}, IWSLT14 German-English \cite{cettolo2014report}, and PersonaChat \cite{zhang2018personalizing}. This evaluation encompasses classification, translation, and generation tasks. Additionally, \citet{jang2022knowledge} conduct experiments on reasoning datasets, such as Hellaswag \cite{zellers2019hellaswag} and MathQA \cite{amini-etal-2019-mathqa}, to assess language models' capabilities before and after the unlearning process.

\paragraph{Evaluation Metrics.} We mainly discuss the unlearning evaluation in natural language field here.  \citet{patil2023can} assess unlearning performance using $\Delta$-Acc, comparing accuracy scores before and after data point deletion on CounterFact \cite{meng2022locating} and zsRE \cite{levy2017zero}. \citet{jang2022knowledge} introduce Extraction Likelihood (EL) and leverage Memorization Accuracy (MA) from \cite{tirumala2022memorization} to quantitatively measure information leakage during extraction attacks. Additionally, \citet{wang-etal-2023-kga} evaluate unlearning through Jensen–Shannon Divergence (JSD), Language Model Probability Distance (LPD), and Proportion of instances with Decreased Language Model Probability (PDLP). It is noteworthy that this method heavily relies on retraining models to assess changes, which can be prohibitively costly, especially in the case of large language models.

\section{Selective Unlearning For LMs}
\subsection{Our Unlearning Methodology}

\paragraph{Problem Formulation} By following \citet{wang-etal-2023-kga}, we formulate machine unlearning as follows. Given a training set $D \in \mathcal{Z}^*$, The process of training a model on data set $D$ is denoted by a function $A:\mathcal{Z}^* \rightarrow \mathcal{H}$, where $Z$ is the space of data instances and $H$ is the hypothesis space of models. Then the trained model can be denoted as $A(D)$. For general machine unlearning, the unlearning mechanism is denoted as a function $U$, which takes a training dataset $D \in \mathcal{Z}^*$ (optional), a forget set $D_f \subset D$ (containing data to be unlearned) and a model $A(D)$ as input, and returns an unlearned model $U(D, D_f, A(D)) \in \mathcal{H}$. 

For our selective unlearning, for each instance in the forget set $D_f$, we define a forget span set that explicitly refers to the sequence spans that need to be unlearned. Formally, for any $x=(x_1, x_2, \ldots, x_T) \in D_f$, the forget span set is denoted as $s^x = (s_1, s_2, \ldots, s_m)$, where $m$ is the number of the forget spans and $s_i$ represents a continuous sub-sequence of $x$, i.e., $s_i = (x_{j_i}, x_{{j_i}+1}, \ldots, x_{{j_i}+|s_i|-1})$, where ${j_i}$ indicates the original index in $x$ of the first token of $s_i$.


\paragraph{Our \shortname Unlearning Method}
Different from \citet{jang2022knowledge} which simply negates the original training objective of minimizing the negative log-likelihood of token sequences as an unlearning method, we propose a selective unlearning method called \shortname. \shortname forgets knowledge in a fine-grained manner, aiming to forget specific sequence spans instead of instance-level forgetting. The motivation mainly comes from the fact that instance-level forgetting cannot handle the adversarial attack mentioned in \Cref{fig:intro} and selective unlearning can have less impact on the performance of the original language model. The learning objective of \shortname is to \textit{minimize} the following loss function for all $x \in D_f$:
\begin{equation}
\label{eq:training_objective}
    \mathcal{L}_{UL}(A(D), x) =  \sum_{s_i \in s^x} \sum_{t=j_i}^{j_i+|s_i|-1} \log(p_{\theta}(x_t | x_{<t}))
\end{equation}
where \( x_{<t} \) denotes the sequence preceding the index $t$,
and \( j_i \) is the start index of subsequence \( s_i \) in original sequence $x$. \( p_{\theta}(x_t | x_{<t}) \) denotes the conditional probability of predicting the next token when given prefix \( x_{t} \) to an LM \( A(D) \) with parameters \( \theta \). 
With the learning objective, the model is pushed to only unlearn the forget spans in forget span set $s^x$ without affecting other general knowledge.
\begin{figure*}[t]
\centering
\includegraphics[width=0.8\linewidth]{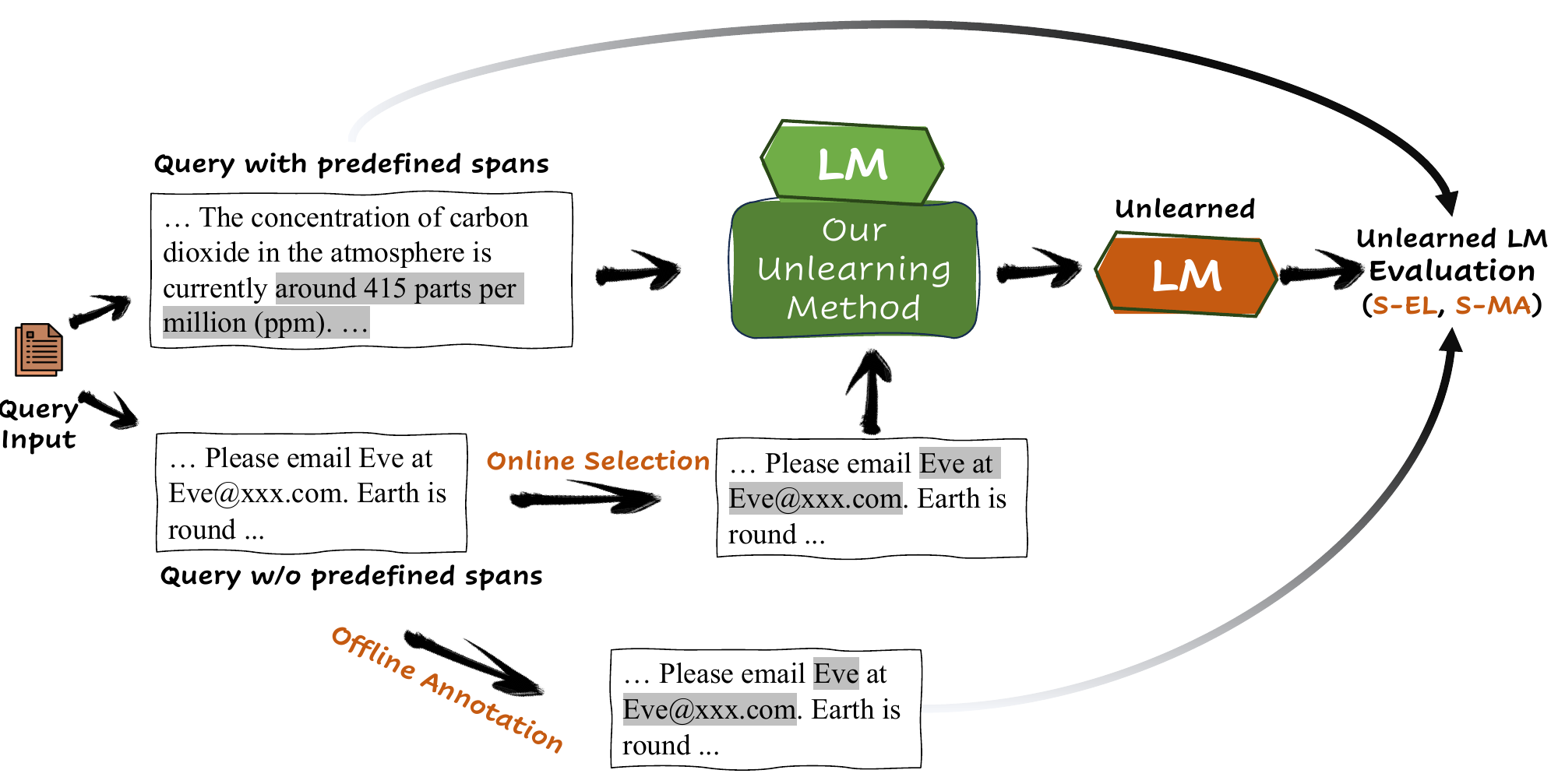}
\vskip -0.5em
\caption{\label{fig:workflow} Workflow of \shortname: Queries with predefined spans (either sensitive or within other definitions) can be inputted directly into \shortname. For queries without predefined spans, we conduct online selection before feeding them to \shortname.}
\vskip -0.5em
\end{figure*}
\subsection{How to Determine the Forgetting Span?}\label{ssec:model:forgetting_span}
It's essential to note that, the forgetting span could be tailored based on practical needs (e.g., for spans with toxic words) in some scenarios. 
However, a more common scenario is that the forget span set is not predefined, and we must develop a span selection/annotation mechanism to achieve the goal of selective unlearning. We propose a comprehensive span selection mechanism designed to adapt to various stages of machine unlearning, specifically addressing concerns related to privacy leakage. In the following sections, we provide details about our forget span selection/annotation design.

\paragraph{Online Selection} When conducting unlearning based on our proposed learning objective (i.e., Eq. \ref{eq:training_objective}) without providing forget spans but for general goal of forgetting (i.e., reducing the privacy leakage risk of $D_f$), we propose an automatic and efficient target span online selection method. We assume that content spans with privacy leakage risks are most likely not common knowledge (i.e. the trained model should not have seen them frequently during training). This means that they are probably with low language probability in language models (a higher perplexity score accordingly). Based on the assumption, we thereby define the following online selection process:
\begin{equation}
    \text{{Select}}(x, \alpha) = \left\{x_t \mid \log(p^{'}_{\theta}(x_t | x_{<t}) < \alpha \right\}
\end{equation}

\noindent where $p^{'}_{\theta}(\cdot)$ indicates the language probability of original model (before unlearning) and $\alpha$ should be a predefined language probability threshold or the average log-probability of the tokens in $x$ (i.e. $\alpha = \sum_{x_t \in x} \log(p^{'}_{\theta}(x_t | x_{<t})$).
We also include tokens that are located between two closely positioned tokens (for any $x_i$ and $x_j$, we consider them close if $|i-j| \leq 2$). This decision is based on our preliminary observation that, in some cases, a complete word or phrase may exhibit a higher language probability in the middle tokens. Then we aggregate all the selectively adjacent tokens into spans, forming $s^x = (s_1, s_2, ..., s_m)$. 
The efficient selection of forget spans described above can be conducted during the training of the language model, thereby facilitating our proposed selective unlearning. As this selection is based on our aforementioned assumption, we will empirically discuss its rationality in the Experimental Results Section.

\paragraph{Offline Annotation} 
Online selection provides a rapid and cost-effective solution, but it may lack the precision required for identifying genuinely privacy-sensitive tokens. To address this limitation, we propose an offline forget spans annotation method that ensures greater accuracy. 
The annotation workflow and the original forget set $D_f$ together with the annotated forget span sets could serve as a valuable resource to evaluate unlearning methods' efficiency in forgetting target knowledge. In the following, we describe our bi-directional verified LLM-based annotation process.
 
(i) Forward Span Annotation. We leverage the strong in-context learning capability of large language models to facilitate forget span annotation. As previously mentioned, this work primarily focuses on the general purpose of unlearning, i.e., forgetting information with high risks of leaking private information. To achieve this, we instruct large language models (e.g., ChatGPT) to annotate sensitive spans in a given text using a few-shot setting. The detailed prompt is provided in the Appendix. Following this prompt, the large language models generate sensitive spans\footnote{An exact token matching-based post-processing is applied to ensure that the generated spans match the original text input.} for each instance $x$ in the forget set $D_f$. The entire process can be summarized as follows:
\begin{equation}
    \check{s}^x \leftarrow \mathcal{F}(x, \mathcal{D})
\end{equation}
\noindent where $\mathcal{D}$ denotes the examples used in few-show learning.



(ii) Backward Verification. Previous research \cite{ling2023deductive,shinn2023reflexion} has demonstrated double-checking the reasoning results generated by LLMs can produce significantly more reliable content. In this work, we employ a backward verification mechanism. Instead of merely prompting LLMs to validate the accuracy of previously generated sensitive spans, we instruct them to independently assess and score the generated spans on a scale of \{0, 1, 2\} without providing the entire sequence (i.e., $x$). Subsequently, we filter out spans with low scores (i.e., score 0).

\subsection{Evaluation of Unlearning}
\label{sec:method:eval}
In this subsection, we present two newly proposed quantitative evaluation metrics designed for assessing machine unlearning with a specific focus on handling sensitive information: Sensitive Extraction Likelihood (S-EL) and Sensitive Memorization Accuracy (S-MA). 

\paragraph{Sensitive Extraction Likelihood}
Before introducing S-EL, we provide a succinct definition of string overlap ($\textsc{Ovl}$) (following \citet{jang2022knowledge}) and our proposed overlap with consideration for sensitive information ($\textsc{S-Ovl}$) given two token sequences $a$ and $b$:
\begin{equation}\label{eq:ovl}
 \textsc{Ovl}_n(a, b) = \dfrac{\sum_{c\in n\text{-grams}(a)} \mathds{1}{\{c \subseteq b\}}}{| n\text{-grams}(a)|}
\end{equation}
\begin{equation}\footnotesize\label{eq:s-ovl}
 \textsc{S-Ovl}_n(a, b) = \dfrac{\sum_{c\in n\text{-grams}(a)} \mathds{1}{\{(c \cap s^b) \neq \varnothing \land (c \cap s^b) \subseteq b\}}}{\sum_{c\in n\text{-grams}(a)} \mathds{1}{\{(c \cap s^b) \neq \varnothing\}}}
\end{equation}

\noindent where $\mathds{1}{\{\cdot\}}$ denotes an indicator function that returns 1 if the evaluated condition is true; otherwise 0. $n\text{-grams}()$ returns the list of n-grams in the given token sequence, and $s^b$ represents the sensitive span set in $b$ that needs to be unlearned. 
$c \subseteq b$ indicates that $c$ is the subsequence of $b$. $c \cap s^b$ returns the common subsequence of $c$ and any span in $s^b$.
Notably, the distinction between \textsc{S-Ovl} and \textsc{Ovl} lies in the fact that \textsc{S-Ovl} exclusively considers n-grams containing sensitive spans.

Building upon \textsc{S-Ovl}, we propose S-EL, which assesses the likelihood of a language model $f_{\theta}$ generating sensitive information when prompted with a prefix in $x$, in the basis of n-gram. 

\begin{equation}
\textsc{S-EL}_n(x) = \dfrac{\sum_{t=1}^{T-n} \textsc{S-Ovl}_n(f_{\theta}(x_{<t}),x_{\geq t})}{T-n}
\end{equation}
\noindent where $f_{\theta}(x_{<t})$ denotes the suffix generated by the language model $f_{\theta}$ given the prefix $x_{<t}$.

\paragraph{Sensitive Memorization Accuracy}
S-MA quantifies the model's accuracy in predicting the next token given a prefix following \citet{tirumala2022memorization}, but only considering the sensitive information. We define it as follows.
\begin{equation}
\textsc{S-MA}(x) = \dfrac{\sum_{t=1}^{T-1} \mathds{1}{\{\text{
argmax} ( p_{\theta}(\cdot|x_{<t})) = x_t \land x_t \in s^x \}}}{\sum_{t=1}^{T-1} \mathds{1}{\{ x_t \in s^x \}}}
\end{equation}

\noindent where $\text{
argmax}(p_{\theta}(\cdot|x_{<t}))$ denotes the most probable next token predicted by the language model. $T$ is the total number of tokens in the sequence $x$ and $x_t \in s^x$ indicates that token $x_t$ is contained in any sensitive span of span set $s^x$. 

Both S-EL and S-MA provide a robust framework for evaluating the efficacy of machine unlearning, particularly in contexts where the handling of sensitive information is a critical concern. 
For comparison, we also denote the extraction likelihood and memorization accuracy without considering whether the tokens are sensitive or not as EL$_{n}$ \cite{jang2022knowledge} and MA \cite{tirumala2022memorization} and display their results together with the two proposed metrics.

\section{Experimental Setup}
\begin{table*}[t]
    \setlength{\tabcolsep}{2.1mm}
    \newcommand{\tabincell}[2]{\begin{tabular}{@{}#1@{}}#2\end{tabular}}
    \begin{center}
        \resizebox{0.75\linewidth}{!}{
            \begin{tabular}{ll|cccc|c|cc|c}
                \toprule
                \multirow{2}{*}{Models} & & \multicolumn{4}{c}{\tabincell{c}{Forget Evaluation}} & \multicolumn{1}{c}{\tabincell{c}{Avg. 8 Cla.}}& \multicolumn{2}{c}{\tabincell{c}{Avg. 4 Dia.}} & Epoch \\
                \cmidrule(lr){3-6}\cmidrule(lr){7-7}\cmidrule(lr){8-9}\cmidrule(lr){10-10}
                & & EL$_{10}$ & MA & S-EL$_{10}$($\downarrow$) & S-MA($\downarrow$) & Acc($\uparrow$) & F1($\uparrow$)  & PPL($\downarrow$) &\#  \\
                \midrule
                \rowcolor[rgb]{0.93,0.93,0.93}
                \multirow{5}{*}{\tabincell{l}{GPT-Neo \\ 125M}} & Original &31.5 &76.9 &22.0 &62.4  &43.7 &7.4 &43.6 & -\\
                \cmidrule(lr){2-10}
                & \textsc{Kul} &0.2 &26.8 & 1.3 &22.9 & 40.9 &1.1 &539.1 &15.6 \\
                
                & \shortname &1.0 &29.5 & \textbf{0.3} &\textbf{16.1} &\textbf{41.0} &\textbf{7.8} &\textbf{179.8} & 16.2  \\
                \midrule
                \rowcolor[rgb]{0.93,0.93,0.93}
                \multirow{5}{*}{\tabincell{l}{GPT-Neo \\ 1.3B}} & Original &60.4 &91.1 &44.6 &81.6  &48.6 &12.3 &26.0 & - \\
                \cmidrule(lr){2-10}
                & \textsc{Kul} &0.3 &27.2 &\textbf{1.4} &23.9  &\textbf{48.2} &10.4 &32.4 & 9.2  \\
                & \shortname &0.6 &29.4 &1.9 &\textbf{23.6} & 48.0 &\textbf{11.4} &\textbf{28.3} &8.8 \\
                \midrule
                \rowcolor[rgb]{0.93,0.93,0.93}
                \multirow{5}{*}{\tabincell{l}{GPT-Neo \\ 2.7B}} & Original &66.6 &93.1 &49.5 &82.5 &50.7 &12.3 &23.3 & - \\
                \cmidrule(lr){2-10}
                & \textsc{Kul} &0.5 &29.8 &1.7 &23.0  &49.9 &10.1 &30.1 &8.6 \\
                & \shortname &0.3 &18.6 &\textbf{1.0} &\textbf{14.6}  &\textbf{50.2} &\textbf{10.9} &\textbf{26.7} &8.8  \\
                \midrule
                \rowcolor[rgb]{0.93,0.93,0.93}
                \multirow{5}{*}{\tabincell{l}{LlaMa2 \\ 7B}} & Original &25.7 &73.2 &19.9 &58.2 &56.9 &13.2  &9.0 & -\\
                \cmidrule(lr){2-10}
                & \textsc{Kul} &0.7 &29.1 &1.5 &22.3 &52.1 &9.7  & 21.0 & 7.1\\
                & \shortname &0.6 &13.9 &\textbf{1.2} &\textbf{12.3} &\textbf{53.6} &\textbf{10.4}  & \textbf{16.3} & 6.8\\
                \midrule
                \rowcolor[rgb]{0.93,0.93,0.93}
                \multirow{5}{*}{\tabincell{l}{Mistral \\ 7B}} & Original &22.9 &75.9 &19.0 &63.9 &62.5 &14.1  & 9.1 & -\\
                \cmidrule(lr){2-10}
                & \textsc{Kul} &0.5 &27.9 &2.1 &23.6 &53.3 &13.3  &19.8 & 7.0\\
                & \shortname &0.4 &19.1 &\textbf{1.3} & \textbf{20.9} &\textbf{55.7} &\textbf{12.8}  &\textbf{17.0} & 7.4\\

                \bottomrule
            \end{tabular}
        }
    \end{center}
    \caption{\label{tab:main_results} Comparison results (in \%) on forget set ($d$=32), 8 classification datasets and 4 dialogue datasets. The best performance in comparable columns are highlighted in \textbf{bold}. ``Avg.'' denotes average scores.}
\end{table*}

\subsection{Datasets and Evaluation Metrics}
As we focus on unlearning of language models, we start from pretrained language models, then we do unlearning on forget dataset $D_f$ and test the models (original LMs and unlearned LMs) on test set $D_t$. 
\paragraph{Forget Dataset $D_f$.} 
In order to ensure a fair comparison with \citet{jang2022knowledge} and to assess the privacy risk associated with language models, we employ the identical samples as those disclosed by \citet{jang2022knowledge}. The forget set is sourced from the Training Data Extraction Challenge\footnote{\url{https://github.com/google-research/lm-extraction-benchmark}}, comprising 15,000 examples, each consisting of 200-token sequences from various domains of Pile corpora.

\paragraph{Evaluation Datasets $D_t$.}
To evaluate general language modeling capabilities, 
we employ 8 classification tasks (i.e., Hellaswag~\citep{zellers2019hellaswag} to measure linguistic reasoning abilities, Winogrande~\citep{sakaguchi2021winogrande}, and COPA~\citep{gordon-etal-2012-semeval} to measure commonsense reasoning abilities, and ARC-Easy~\citep{Clark2018ThinkYH}, ARC-Challenge~\citep{Clark2018ThinkYH}, Piqa~\citep{bisk2020piqa}, MathQA~\citep{amini-etal-2019-mathqa}, PubmedQA~\citep{jin2019pubmedqa} benchmarks to measure scientific reasoning abilities) and 4 dialogue tasks (Wizard of Wikipedia~\citep{dinan2018wizard}, Empathetic Dialogues~\citep{rashkin-etal-2019-towards}, Blended Skill Talk~\citep{smith-etal-2020-put}, and Wizard of Internet~\citep{komeili-etal-2022-internet}). These test sets are used to assess whether the general capabilities of language models are affected after unlearning. 

\paragraph{Evaluation Metrics.} We assess the performance of the methods from two perspectives: the effectiveness of unlearning and the maintenance of general performance. To evaluate the unlearning effectiveness on the forget set, we employ our proposed metrics S-EL$_{n}$ and S-MA, as well as EL$_{n}$ and MA, introduced in Section~\ref{sec:method:eval}. As for the evaluation of maintaining performance, we use accuracy for the classification test sets and F1 and Perplexity (PPL) scores for dialogue tasks.

\subsection{Baselines and Implementation Details}
\label{ssec:baseline_threshold}

\paragraph{Baselines and Forgetting Threshold.} Our primary comparative analysis involves contrasting our \shortname with the approach proposed by \citet{jang2022knowledge} (referred to as KUL). 
We follow their specified forgetting thresholds, based on $\textsc{EL}_{10}$ and MA representing the extraction likelihood and memorization accuracy of regular data. Importantly, these thresholds serve as empirical indicators of the targeted forgetting objectives for unlearning methods.
For fair comparison, the results of \shortname reported are all achieved with our \textit{online selection} method. We denote our results with \textit{offline annotation} as \textit{Oracle} results and are compared in Table~\ref{tab:oracle_training}.

\paragraph{Models and Training Details.} The unlearning experiments are conducted on the pre-trained language model GPT-Neo series (125M, 1.3B, and 2.7B), Llama2-7B and Mistral-7B. 
The learning rate for training is set to $5 \times 10^{-5}$, based on the selection from $[2 \times 10^{-5}, 5 \times 10^{-5}, 1 \times 10^{-4}]$. 
The variable denoting the number of forgetting instances, represented as $d$, is examined across the values $d=1, 2, 4, 8, 16, 32, 64, 128$. Unless otherwise specified, the reported results in this paper are based on the $d=32$ setting.
We adapt the global batch size during training to be the same as $d$, the number of forgetting instances, following \citet{jang2022knowledge}.
Each setting is run 5 times and the reported results are the average of 5 different runs.
All the models are trained with a single Nvidia GeForce RTX 3090.

\section{Experimental Results}
\subsection{Main Results}\label{ssec:main_results}
The main comparative results of unlearning methods are presented in \Cref{tab:main_results}, we have the following observations:

$\bullet$~\textit{Our \shortname generally exhibits superior effectiveness in unlearning sensitive information compared to the baseline.} As explained in Experimental Setup Section, we set the forgetting threshold consistent for both \shortname and \textsc{Kul}. This can be reflected by the results reported in \Cref{tab:main_results} that they exhibit the same levels of EL$_{10}$ and MA. In this context, our \shortname demonstrates better scores in S-EL$_{10}$ and S-MA compared to \textsc{Kul}. Given that EL$_{10}$ and MA reflect general information leakage risk, and S-EL$_{10}$ and S-MA scores emphasize the risk of sensitive information leakage, these results indicate that our \shortname unlearning method offers better protection against sensitive information leakage.

$\bullet$~\textit{Our \shortname demonstrates comparable results on classification datasets when compared to the \textsc{Kul} method, but it exhibits significantly better performance on dialogue datasets.} When examining the average accuracy scores on classification tasks, we observe that \shortname maintains an accuracy difference of $0.1-0.3\%$ compared to the \textsc{Kul} method across three different GPT-Neo language model backbones (125M, 1.3B, and 2.7B). In contrast, when considering the average F1 and PPL scores on dialogue datasets, \shortname significantly outperforms \textsc{Kul}. This improvement may stem from that our fine-grained unlearning approach minimally impacts the generation performance of the language model, unlike \textsc{Kul} unlearning method that completely negate the loss function.

$\bullet$~\textit{Our unlearning target does not result in longer training epochs.} Upon examining the average number of epochs required to reach the forgetting threshold, we observe that the average epochs for \shortname and \textsc{Kul} are similar. This suggests that our approach, which involves partially negating the loss function of pretraining, does not prolong the unlearning process compared to the fully reversed loss function employed by \textsc{Kul}.
\begin{figure}[t]
\centering
\subfigure[Scores over Epoch] {\label{sfig:epoch_1.3b_mael}
\includegraphics[width=0.45\linewidth]{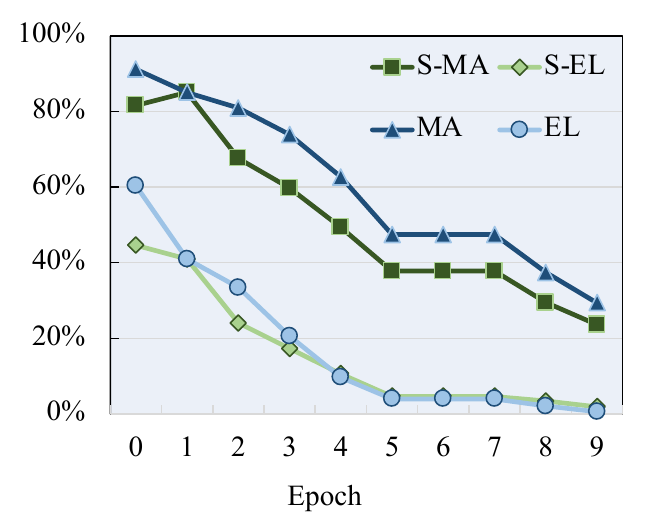}
}
\subfigure[Scores over Epoch] {\label{sfig:epoch_1.3b_accf1}
\includegraphics[width=0.45\linewidth]{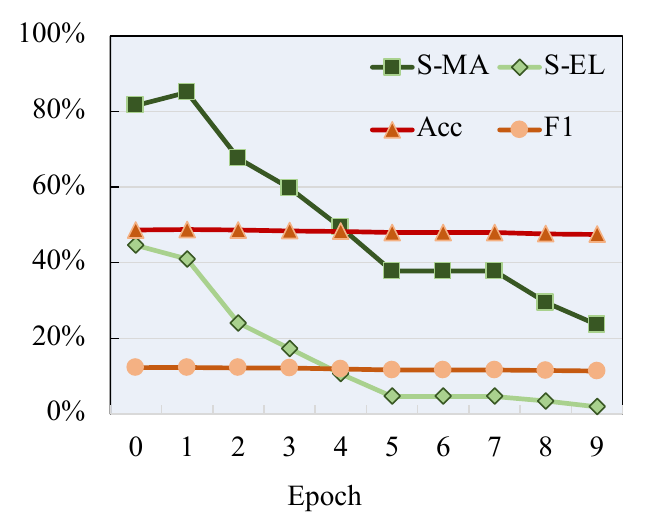}
}
\caption{\label{fig:epoch_1.3b} Unlearning GPT-Neo 1.3B: (a) S-MA, S-EL, MA, EL, and (b) Accuracy, F1 Scores over epochs.
}
\vskip -1em
\end{figure}

$\bullet$~\textit{Smaller language models exhibit less stability in maintaining performance.} When examining the unlearning results across different language models, ranging from small to large (125M to 2.7B), it becomes apparent that the 125M GPT-Neo model experiences the most significant performance drop when forgetting $d=32$ instances.

\subsection{Effectiveness Analysis of \shortname}

\paragraph{Performance Over Epochs.} We illustrate the evolution of scores across training epochs, focusing on Sensitive Extraction Likelihood (S-EL), Sensitive Memorization Accuracy (S-MA), EL, and MA scores in \Cref{sfig:epoch_1.3b_mael}. Additionally, \Cref{sfig:epoch_1.3b_accf1} displays the trends in S-EL, S-MA, average accuracy for classification datasets, and average F1 for dialogue datasets. 
In \Cref{sfig:epoch_1.3b_mael}, the evaluation results of unlearning, with and without consideration of sensitive spans, demonstrate general consistency (i.e., S-MA and S-EL compared to MA and EL). 
Observing the results in \Cref{sfig:epoch_1.3b_accf1}, we note that as unlearning progresses (i.e., S-MA and S-EL degrade), the impact on both classification (Accuracy score) and dialogue (F1 score) performance is minimal. This observation underscores the effectiveness of our unlearning methods.

\begin{figure}[t]
\centering
\subfigure[Accuracy over $d$] {\label{sfig:n_acc}
\includegraphics[width=0.45\linewidth]{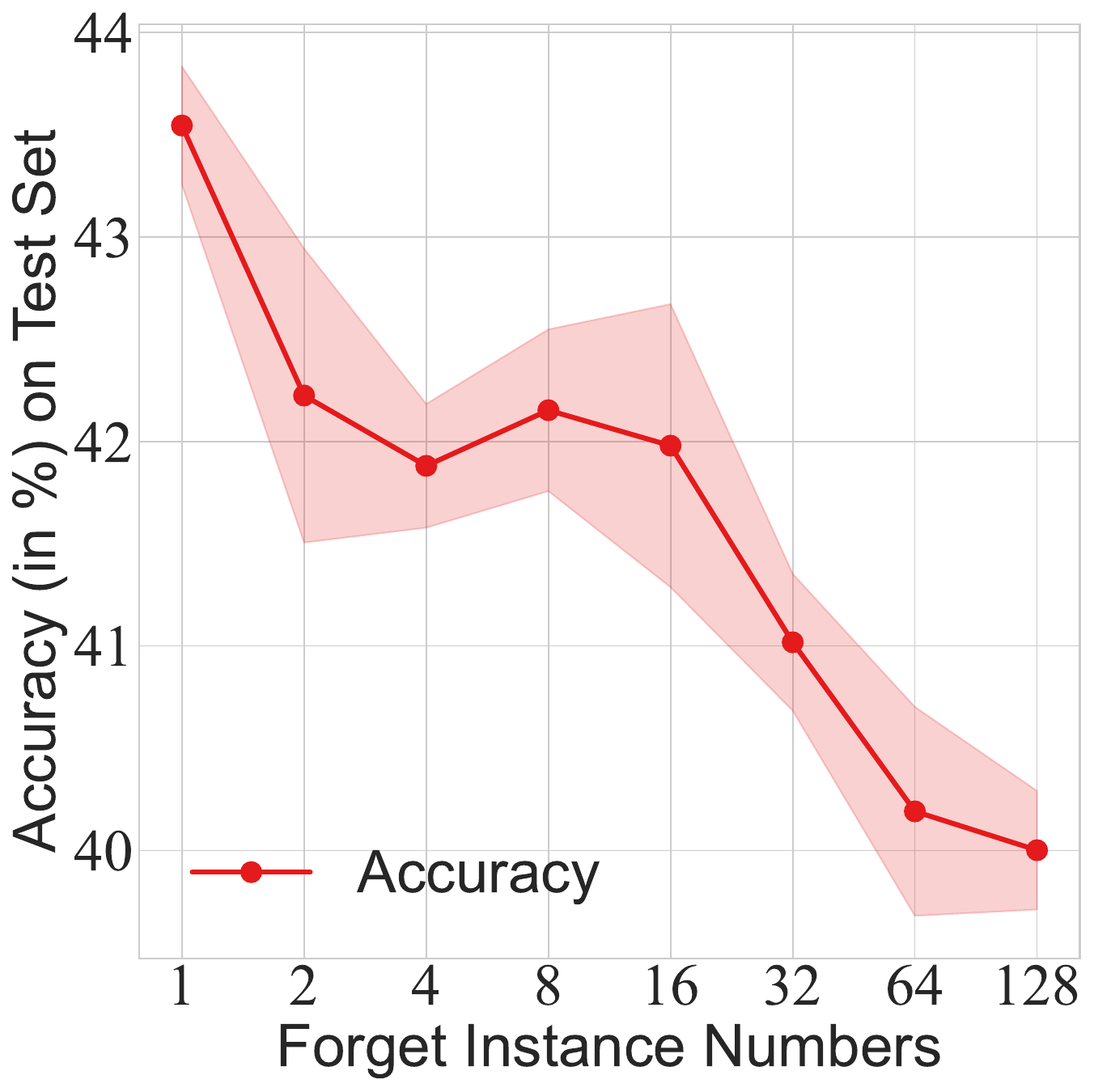}
}
\subfigure[F1 over $d$] {\label{sfig:n_f1}
\includegraphics[width=0.45\linewidth]{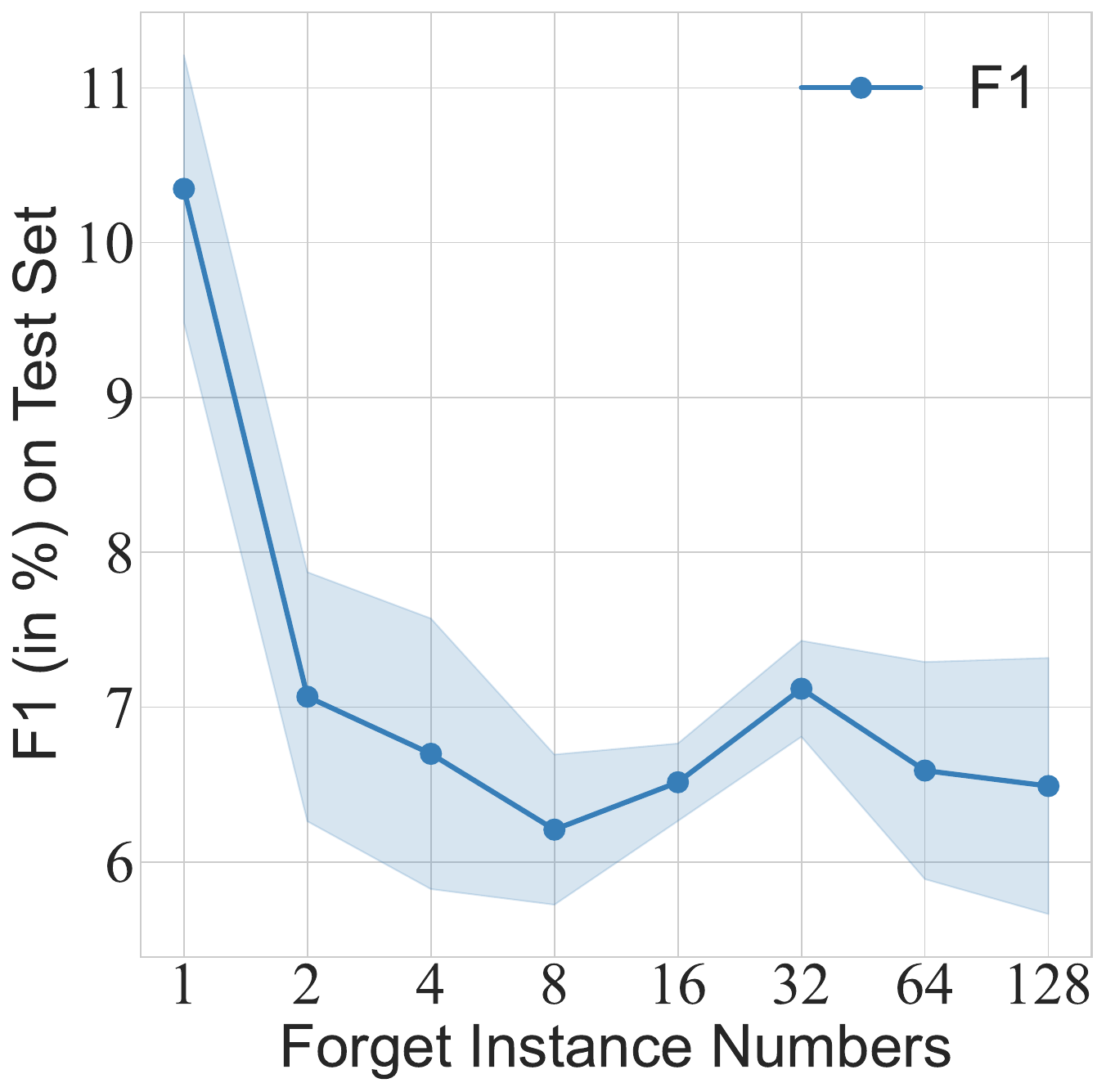}
}
\caption{\label{fig:n_acc_f1} Unlearning GPT-Neo 125M: (a) Accuracy and (b) F1 Scores with varying $d$ values.
}
\vskip -1em
\end{figure}
\paragraph{Performance with Varying $d$.} The main results presented in \Cref{tab:main_results} are derived from a forgetting set with a size of $32$. Here, we investigate the impact of unlearning based on different numbers of forgetting instances. The results for unlearning GPT-Neo 125M are shown in \Cref{fig:n_acc_f1}. 
Analyzing the trend in classification accuracy shown in \Cref{sfig:n_acc}, we observe that, while the overall performance drop is not substantial, there is a noticeable degrading trend as the number of forgetting instances increases. However, when assessing generation performance using the F1 score (\Cref{sfig:n_f1}), we note that as $n$ increases, the F1 scores remain more stable than the classification performance. \Cref{sfig:n_f1} highlights the superiority of \shortname in maintaining generation performance, consistent with the observations discussed in Main Results Subsection based on \Cref{tab:main_results}.

\definecolor{mydarkgreen}{RGB}{0,100,0}
\definecolor{mylightblue}{RGB}{173,216,230}
\begin{table}[t]
\setlength{\tabcolsep}{1mm}\small
\begin{center}
\resizebox{\linewidth}{!}{
\begin{tabular}{l|m{6cm}}
\toprule[1.0pt]
Prefix &Trump is the 45th \\
\midrule
Original & president of the United States. He is the first president to be elected in ... \\
\midrule
\textsc{Kul} & Nation to Target Black Voters Ahead of Election Day ...\\
\midrule
\shortname & president of the United States. The current president of the United States ... \\
\bottomrule[1.0pt]
\end{tabular}
}
\end{center}
\caption{\label{tab:attack_results} Given a prefix, the suffixes generated by the original GPT-Neo 1.3B model, \textsc{Kul} and \shortname unlearned. 
}
\vskip -1em
\end{table}
 
\definecolor{mydarkgreen}{RGB}{0,100,0}
\definecolor{mylightblue}{RGB}{173,216,230}
\begin{table*}[t]
\setlength{\tabcolsep}{1mm}\small
\begin{center}
\begin{tabular}{p{1.7cm}|m{15.5cm}}
\toprule[1.0pt]

\textbf{Method} & \textbf{Span Annotation} \\

\midrule[0.5pt]
\midrule
\textbf{NER} & XXXX@XXXmail.com>.* * Permission is hereby granted, free of charge, to any person obtaining a copy* of this \colorbox{mylightblue}{software} and associated documentation files (the \colorbox{mylightblue}{Software}), to deal* in the \colorbox{mylightblue}{Software} without restriction, including without limitation the rights* to use, copy, modify, merge, publish, distribute, sublicense, and/or sell* copies of the \colorbox{mylightblue}{Software}, ... \\
\midrule
\textbf{Online Selection} & \colorbox{mylightblue}{XXXX@XXXmail.com>.*} * Permission is hereby granted, free of charge, to any person obtaining a \colorbox{mylightblue}{copy*} of this software and associated documentation files (the Software), to deal* in the Software without restriction, including without limitation the rights* to use, copy, modify, merge, publish, distribute, sublicense, and/or sell* copies of the Software, ... \\
\midrule
\textbf{Offline Annotation} & \colorbox{mylightblue}{XXXX@XXXmail.com}>.* * Permission is hereby granted, free of charge, to any person obtaining a copy* of this software and associated documentation files (the Software), to deal* in the Software without restriction, including without limitation the rights* to use, copy, modify, merge, publish, distribute, sublicense, and/or sell* copies of the Software, ... \\

\bottomrule[1.0pt]
\end{tabular}
\end{center}
\vskip -1em
\caption{\label{tab:case_study} Comparing annotated \colorbox{mylightblue}{spans} of NER, Online Selection, and LLM-based Offline Annotation.
}
\vskip -1em
\end{table*}
\paragraph{Stability under Adversarial Attacks.} As illustrated in \Cref{fig:intro}, attackers may inject common sense or general knowledge into their original information and subsequently request unlearning. Under such attacks, full loss negation-based method (i.e., \textsc{Kul} \cite{jang2022knowledge}), may inadvertently forget the certain knowledge during unlearning. In contrast, our method demonstrates a capacity to withstand such attacks due to our selective unlearning. To simulate this adversarial scenario, we add additional common knowledge (e.g., ``\textit{Trump is the 45th president of the United States.}'') — information easily generated by the original language model — into every instances in forget dataset $D_f$ and then conduct unlearning. We showcase the continuation results of models (with greedy decoding) when given the prefix ``\textit{Trump is the 45th}'' in \Cref{tab:attack_results}. The results in \Cref{tab:attack_results} demonstrate that \textsc{Kul} is unable to continue the commonsense sentence after unlearning, while our \shortname method can handle such attacks. This underscores the superiority and robustness of our unlearning methods.

\begin{table}[t]
    \setlength{\tabcolsep}{2.1mm}\small
    \newcommand{\tabincell}[2]{\begin{tabular}{@{}#1@{}}#2\end{tabular}}
    \begin{center}
        \resizebox{\linewidth}{!}{
            \begin{tabular}{l|cc|c|cc|c}
                \toprule
                \multirow{2}{*}{Models} & \multicolumn{2}{c}{\tabincell{c}{Forget Evaluation}} & \multicolumn{1}{c}{\tabincell{c}{Avg. 8 Cla.}}& \multicolumn{2}{c}{\tabincell{c}{Avg. 4 Dia.}} & Epoch \\
                \cmidrule(lr){2-3}\cmidrule(lr){4-4}\cmidrule(lr){5-6}\cmidrule(lr){7-7}
                & S-EL$_{10}$ & S-MA & Acc($\uparrow$) & F1($\uparrow$)  & PPL($\downarrow$) &\#  \\
                \midrule
                
                \shortname  & 0.3 &16.1&41.0 &\textbf{7.8} &179.8 & 16.2  \\
                \cmidrule(lr){2-7}
                \shortname$^{O}$ &0.3 &14.6 & \textbf{41.5} &\textbf{7.8} &\textbf{102.4} & 13.0  \\
                \bottomrule
            \end{tabular}
        }
    \end{center}
    \vskip -1em
    \caption{\label{tab:oracle_training} Results (in \%) for \shortname (trained on online selected data) and \shortname$^{Oracle}$ (trained on offline annotated data).}
    \vskip -1em
\end{table}

\begin{table}[t]
    \setlength{\tabcolsep}{2.1mm}\small
    \newcommand{\tabincell}[2]{\begin{tabular}{@{}#1@{}}#2\end{tabular}}
    \begin{center}
        \resizebox{\linewidth}{!}{
            \begin{tabular}{l|ccc}
                \toprule
                Method &  Avg. Span \#  &  Avg. Prop.  & Cover (Human, $\cdot$) \\
                \midrule
                Human & 1.8 & 3.1\% & - \\
                NER & 19.2 & 8.9\% & 5.3\% \\
                Online & 8.6 & 13.7\% & 72.0\% \\
                Offline & 4.5 & 9.5\% & 75.1\% \\
                \bottomrule
            \end{tabular}
        }
    \end{center}
    \vspace{-1em}
    \caption{\label{tab:annotation_statistic} Statistics of sensitive spans annotated by various methods. Avg. Span \# refers to the average spans per instance, Avg. Prop. refers to the proportion of sensitive spans in the sequence. Cover (Human, $\cdot$) indicates the percentage of human-annotated spans covered by other annotations.}
    \vspace{-1em}
\end{table}

\paragraph{Oracle Sensitive Span Based Unlearning.} It is noteworthy that our approach based on online selection emphasizes avoiding reliance on pre-determined sensitive or forgetting spans to produce a more general and cost-efficient solution. 
However, this inevitably introduces some inaccurately detected spans when evaluating with S-EL and S-MA based on ChatGPT-annotated spans. 
Therefore, we further explore unlearning based on offline annotated sensitive spans (denoted as \textsc{SeUL}$^{Oracle}$) to assess the magnitude of this discrepancy. 
We stop the training of \textsc{SeUL}$^{Oracle}$ when its S-EL$_{10}$ S-MA scores are less or equal to \shortname, and then report the corresponding accuracy and F1 scores in  \Cref{tab:oracle_training}. From the results in \Cref{tab:oracle_training}, we can see that \textsc{SeUL}$^{Oracle}$ shows a better capability in maintaining the performance of language model, thanks to its accurate annotation and less training epochs. 

\subsection{Reliability of Sensitive Span Annotation}\label{ssec:reliability_annotation}
\shortname's training and evaluation rely on automatic online or offline annotated sensitive spans. To ascertain the reliability of the annotation, we compare our automatic annotation with human annotation and Named Entity Recognition (NER) based annotation. Specifically, we invite two annotators to do sensitivity annotation on 50 instances to be forgotten and then get 50 annotated instances regarded as ground truth. We also employ NER toolkit to annotate the same 50 instances where extracted entities are regarded as sensitive information. We have the following quantitative and qualitative analyses.

\paragraph{Quantitative Comparison.} We conduct a statistical analysis of annotated sensitive spans derived by these methods and display results in \Cref{tab:annotation_statistic}. The results reveal that human annotators apply a more stringent criterion for sensitive annotation, leading to a smaller average number of sensitive spans per instance. In contrast, NER-based annotation labels more spans, but these spans exhibit a lower overlap with the spans annotated by human annotators. Our online sensitive selection method tends to label more spans while offline annotation achieves the highest matching with human results.

\paragraph{Case Study.} We present a case study using the example in \Cref{tab:case_study} to illustrate the annotation results of NER, Online Selection, and Offline Annotation. As observed, NER primarily extracts potential entities from the sequence, without specifically identifying sensitive information. Both Online Selection and Offline Annotation successfully annotate sensitive information, e.g. the email address (masked to avoid privacy issues). Nevertheless, it's important to note that Online Selection may introduce additional spans that do not accurately represent sensitive information.

\section{Conclusion}
In summary, this paper presents \shortname, a novel selective unlearning method for language models that minimizes negative impacts on model capabilities by focusing on specific sequence spans. It introduces specialized evaluation metrics, S-EL and S-MA, designed to assess the forgetting of sensitive information. The paper also proposes efficient automatic online and offline sensitive span annotation methods to support the overall unlearning framework. Overall, these contributions address concerns regarding the inadvertent retention of personal or sensitive information by neural models.

\section{Acknowledgments}
The research described in this paper is partially supported by HK RGC GRF \#14206324, The National Natural Science Foundation of China (Grant No. 62227808) and Shenzhen Science and Technology Program (Grant No.ZDSYS20210623091809029). 

\bibliography{anthology,custom}

\begin{thebibliography}{35}
\providecommand{\natexlab}[1]{#1}

\bibitem[{Amini et~al.(2019)Amini, Gabriel, Lin, Koncel-Kedziorski, Choi, and Hajishirzi}]{amini-etal-2019-mathqa}
Amini, A.; Gabriel, S.; Lin, S.; Koncel-Kedziorski, R.; Choi, Y.; and Hajishirzi, H. 2019.
\newblock {M}ath{QA}: Towards Interpretable Math Word Problem Solving with Operation-Based Formalisms.
\newblock In \emph{Proceedings of the 2019 Conference of the North {A}merican Chapter of the Association for Computational Linguistics: Human Language Technologies, Volume 1 (Long and Short Papers)}, 2357--2367. Minneapolis, Minnesota: Association for Computational Linguistics.

\bibitem[{Bisk et~al.(2020)Bisk, Zellers, Gao, Choi et~al.}]{bisk2020piqa}
Bisk, Y.; Zellers, R.; Gao, J.; Choi, Y.; et~al. 2020.
\newblock Piqa: Reasoning about physical commonsense in natural language.
\newblock In \emph{Proceedings of the AAAI conference on artificial intelligence}, volume~34, 7432--7439.

\bibitem[{Bourtoule et~al.(2021)Bourtoule, Chandrasekaran, Choquette-Choo, Jia, Travers, Zhang, Lie, and Papernot}]{bourtoule2021machine}
Bourtoule, L.; Chandrasekaran, V.; Choquette-Choo, C.~A.; Jia, H.; Travers, A.; Zhang, B.; Lie, D.; and Papernot, N. 2021.
\newblock Machine unlearning.
\newblock In \emph{2021 IEEE Symposium on Security and Privacy (SP)}, 141--159. IEEE.

\bibitem[{Cao and Yang(2015)}]{cao2015towards}
Cao, Y.; and Yang, J. 2015.
\newblock Towards making systems forget with machine unlearning.
\newblock In \emph{2015 IEEE Symposium on Security and Privacy}, 463--480. IEEE.

\bibitem[{Cettolo et~al.(2014)Cettolo, Niehues, St{\"u}ker, Bentivogli, and Federico}]{cettolo2014report}
Cettolo, M.; Niehues, J.; St{\"u}ker, S.; Bentivogli, L.; and Federico, M. 2014.
\newblock Report on the 11th IWSLT evaluation campaign.
\newblock In \emph{Proceedings of the 11th International Workshop on Spoken Language Translation: Evaluation Campaign}.

\bibitem[{Clark et~al.(2018)Clark, Cowhey, Etzioni, Khot, Sabharwal, Schoenick, and Tafjord}]{Clark2018ThinkYH}
Clark, P.; Cowhey, I.; Etzioni, O.; Khot, T.; Sabharwal, A.; Schoenick, C.; and Tafjord, O. 2018.
\newblock Think you have Solved Question Answering? Try ARC, the AI2 Reasoning Challenge.
\newblock \emph{ArXiv}, abs/1803.05457.

\bibitem[{Dinan et~al.(2019)Dinan, Roller, Shuster, Fan, Auli, and Weston}]{dinan2018wizard}
Dinan, E.; Roller, S.; Shuster, K.; Fan, A.; Auli, M.; and Weston, J. 2019.
\newblock Wizard of Wikipedia: Knowledge-Powered Conversational Agents.
\newblock In \emph{International Conference on Learning Representations}.

\bibitem[{Fan, Lewis, and Dauphin(2018)}]{fan2018hierarchical}
Fan, A.; Lewis, M.; and Dauphin, Y. 2018.
\newblock Hierarchical neural story generation.
\newblock \emph{arXiv preprint arXiv:1805.04833}.

\bibitem[{Ginart et~al.(2019)Ginart, Guan, Valiant, and Zou}]{ginart2019making}
Ginart, A.; Guan, M.; Valiant, G.; and Zou, J.~Y. 2019.
\newblock Making ai forget you: Data deletion in machine learning.
\newblock \emph{Advances in neural information processing systems}, 32.

\bibitem[{Golatkar, Achille, and Soatto(2020{\natexlab{a}})}]{golatkar2020eternal}
Golatkar, A.; Achille, A.; and Soatto, S. 2020{\natexlab{a}}.
\newblock Eternal sunshine of the spotless net: Selective forgetting in deep networks.
\newblock In \emph{Proceedings of the IEEE/CVF Conference on Computer Vision and Pattern Recognition}, 9304--9312.

\bibitem[{Golatkar, Achille, and Soatto(2020{\natexlab{b}})}]{golatkar2020forgetting}
Golatkar, A.; Achille, A.; and Soatto, S. 2020{\natexlab{b}}.
\newblock Forgetting outside the box: Scrubbing deep networks of information accessible from input-output observations.
\newblock In \emph{European Conference on Computer Vision}, 383--398. Springer.

\bibitem[{Gordon, Kozareva, and Roemmele(2012)}]{gordon-etal-2012-semeval}
Gordon, A.; Kozareva, Z.; and Roemmele, M. 2012.
\newblock {S}em{E}val-2012 Task 7: Choice of Plausible Alternatives: An Evaluation of Commonsense Causal Reasoning.
\newblock In \emph{*{SEM} 2012: The First Joint Conference on Lexical and Computational Semantics {--} Volume 1: Proceedings of the main conference and the shared task, and Volume 2: Proceedings of the Sixth International Workshop on Semantic Evaluation ({S}em{E}val 2012)}, 394--398. Montr{\'e}al, Canada: Association for Computational Linguistics.

\bibitem[{Guo et~al.(2019)Guo, Goldstein, Hannun, and Van Der~Maaten}]{guo2019certified}
Guo, C.; Goldstein, T.; Hannun, A.; and Van Der~Maaten, L. 2019.
\newblock Certified data removal from machine learning models.
\newblock \emph{arXiv preprint arXiv:1911.03030}.

\bibitem[{Jang et~al.(2023)Jang, Yoon, Yang, Cha, Lee, Logeswaran, and Seo}]{jang2022knowledge}
Jang, J.; Yoon, D.; Yang, S.; Cha, S.; Lee, M.; Logeswaran, L.; and Seo, M. 2023.
\newblock Knowledge Unlearning for Mitigating Privacy Risks in Language Models.
\newblock In \emph{Proceedings of the 61st Annual Meeting of the Association for Computational Linguistics (Volume 1: Long Papers)}, 14389--14408. Association for Computational Linguistics.

\bibitem[{Jin et~al.(2019)Jin, Dhingra, Liu, Cohen, and Lu}]{jin2019pubmedqa}
Jin, Q.; Dhingra, B.; Liu, Z.; Cohen, W.; and Lu, X. 2019.
\newblock PubMedQA: A Dataset for Biomedical Research Question Answering.
\newblock In \emph{Proceedings of the 2019 Conference on Empirical Methods in Natural Language Processing and the 9th International Joint Conference on Natural Language Processing (EMNLP-IJCNLP)}, 2567--2577.

\bibitem[{Karasuyama and Takeuchi(2009)}]{karasuyama2009multiple}
Karasuyama, M.; and Takeuchi, I. 2009.
\newblock Multiple incremental decremental learning of support vector machines.
\newblock \emph{Advances in neural information processing systems}, 22.

\bibitem[{Koh and Liang(2017)}]{koh2017understanding}
Koh, P.~W.; and Liang, P. 2017.
\newblock Understanding black-box predictions via influence functions.
\newblock In \emph{International conference on machine learning}, 1885--1894. PMLR.

\bibitem[{Komeili, Shuster, and Weston(2022)}]{komeili-etal-2022-internet}
Komeili, M.; Shuster, K.; and Weston, J. 2022.
\newblock {I}nternet-Augmented Dialogue Generation.
\newblock In \emph{Proceedings of the 60th Annual Meeting of the Association for Computational Linguistics (Volume 1: Long Papers)}, 8460--8478. Dublin, Ireland: Association for Computational Linguistics.

\bibitem[{Levy et~al.(2017)Levy, Seo, Choi, and Zettlemoyer}]{levy2017zero}
Levy, O.; Seo, M.; Choi, E.; and Zettlemoyer, L. 2017.
\newblock Zero-shot relation extraction via reading comprehension.
\newblock \emph{arXiv preprint arXiv:1706.04115}.

\bibitem[{Ling et~al.(2023)Ling, Fang, Li, Huang, Lee, Memisevic, and Su}]{ling2023deductive}
Ling, Z.; Fang, Y.; Li, X.; Huang, Z.; Lee, M.; Memisevic, R.; and Su, H. 2023.
\newblock Deductive Verification of Chain-of-Thought Reasoning.
\newblock \emph{arXiv preprint arXiv:2306.03872}.

\bibitem[{Liu et~al.(2021)Liu, Sap, Lu, Swayamdipta, Bhagavatula, Smith, and Choi}]{liu2021dexperts}
Liu, A.; Sap, M.; Lu, X.; Swayamdipta, S.; Bhagavatula, C.; Smith, N.~A.; and Choi, Y. 2021.
\newblock DExperts: Decoding-time controlled text generation with experts and anti-experts.
\newblock \emph{arXiv preprint arXiv:2105.03023}.

\bibitem[{Lu et~al.(2022)Lu, Welleck, Hessel, Jiang, Qin, West, Ammanabrolu, and Choi}]{lu2022quark}
Lu, X.; Welleck, S.; Hessel, J.; Jiang, L.; Qin, L.; West, P.; Ammanabrolu, P.; and Choi, Y. 2022.
\newblock Quark: Controllable text generation with reinforced unlearning.
\newblock \emph{Advances in neural information processing systems}, 35: 27591--27609.

\bibitem[{Mehta et~al.(2022)Mehta, Pal, Singh, and Ravi}]{mehta2022deep}
Mehta, R.; Pal, S.; Singh, V.; and Ravi, S.~N. 2022.
\newblock Deep Unlearning via Randomized Conditionally Independent Hessians.
\newblock In \emph{Proceedings of the IEEE/CVF Conference on Computer Vision and Pattern Recognition}, 10422--10431.

\bibitem[{Meng et~al.(2022)Meng, Bau, Andonian, and Belinkov}]{meng2022locating}
Meng, K.; Bau, D.; Andonian, A.; and Belinkov, Y. 2022.
\newblock Locating and editing factual knowledge in gpt.
\newblock \emph{arXiv preprint arXiv:2202.05262}, 1.

\bibitem[{Patil, Hase, and Bansal(2023)}]{patil2023can}
Patil, V.; Hase, P.; and Bansal, M. 2023.
\newblock Can Sensitive Information Be Deleted From LLMs? Objectives for Defending Against Extraction Attacks.
\newblock \emph{arXiv preprint arXiv:2309.17410}.

\bibitem[{Rashkin et~al.(2019)Rashkin, Smith, Li, and Boureau}]{rashkin-etal-2019-towards}
Rashkin, H.; Smith, E.~M.; Li, M.; and Boureau, Y.-L. 2019.
\newblock Towards Empathetic Open-domain Conversation Models: A New Benchmark and Dataset.
\newblock In \emph{Proceedings of the 57th Annual Meeting of the Association for Computational Linguistics}, 5370--5381. Florence, Italy: Association for Computational Linguistics.

\bibitem[{Romero, Barrio, and Belanche(2007)}]{romero2007incremental}
Romero, E.; Barrio, I.; and Belanche, L. 2007.
\newblock Incremental and decremental learning for linear support vector machines.
\newblock In \emph{International Conference on Artificial Neural Networks}, 209--218. Springer.

\bibitem[{Sakaguchi et~al.(2021)Sakaguchi, Bras, Bhagavatula, and Choi}]{sakaguchi2021winogrande}
Sakaguchi, K.; Bras, R.~L.; Bhagavatula, C.; and Choi, Y. 2021.
\newblock Winogrande: An adversarial winograd schema challenge at scale.
\newblock \emph{Communications of the ACM}, 64(9): 99--106.

\bibitem[{Shinn et~al.(2023)Shinn, Cassano, Gopinath, Narasimhan, and Yao}]{shinn2023reflexion}
Shinn, N.; Cassano, F.; Gopinath, A.; Narasimhan, K.~R.; and Yao, S. 2023.
\newblock Reflexion: Language agents with verbal reinforcement learning.
\newblock In \emph{Thirty-seventh Conference on Neural Information Processing Systems}.

\bibitem[{Smith et~al.(2020)Smith, Williamson, Shuster, Weston, and Boureau}]{smith-etal-2020-put}
Smith, E.~M.; Williamson, M.; Shuster, K.; Weston, J.; and Boureau, Y.-L. 2020.
\newblock Can You Put it All Together: Evaluating Conversational Agents{'} Ability to Blend Skills.
\newblock In \emph{Proceedings of the 58th Annual Meeting of the Association for Computational Linguistics}, 2021--2030. Online: Association for Computational Linguistics.

\bibitem[{Tirumala et~al.(2022)Tirumala, Markosyan, Zettlemoyer, and Aghajanyan}]{tirumala2022memorization}
Tirumala, K.; Markosyan, A.; Zettlemoyer, L.; and Aghajanyan, A. 2022.
\newblock Memorization without overfitting: Analyzing the training dynamics of large language models.
\newblock \emph{Advances in Neural Information Processing Systems}, 35: 38274--38290.

\bibitem[{Tuggener et~al.(2020)Tuggener, Von~D{\"a}niken, Peetz, and Cieliebak}]{tuggener2020ledgar}
Tuggener, D.; Von~D{\"a}niken, P.; Peetz, T.; and Cieliebak, M. 2020.
\newblock LEDGAR: A large-scale multi-label corpus for text classification of legal provisions in contracts.
\newblock In \emph{Proceedings of the Twelfth Language Resources and Evaluation Conference}, 1235--1241.

\bibitem[{Wang et~al.(2023)Wang, Chen, Yuan, Zeng, Wong, and Yin}]{wang-etal-2023-kga}
Wang, L.; Chen, T.; Yuan, W.; Zeng, X.; Wong, K.-F.; and Yin, H. 2023.
\newblock {KGA}: A General Machine Unlearning Framework Based on Knowledge Gap Alignment.
\newblock In \emph{Proceedings of the 61st Annual Meeting of the Association for Computational Linguistics (Volume 1: Long Papers)}, 13264--13276. Toronto, Canada: Association for Computational Linguistics.

\bibitem[{Zellers et~al.(2019)Zellers, Holtzman, Bisk, Farhadi, and Choi}]{zellers2019hellaswag}
Zellers, R.; Holtzman, A.; Bisk, Y.; Farhadi, A.; and Choi, Y. 2019.
\newblock Hellaswag: Can a machine really finish your sentence?
\newblock \emph{arXiv preprint arXiv:1905.07830}.

\bibitem[{Zhang et~al.(2018)Zhang, Dinan, Urbanek, Szlam, Kiela, and Weston}]{zhang2018personalizing}
Zhang, S.; Dinan, E.; Urbanek, J.; Szlam, A.; Kiela, D.; and Weston, J. 2018.
\newblock Personalizing Dialogue Agents: I have a dog, do you have pets too?
\newblock In \emph{Proceedings of the 56th Annual Meeting of the Association for Computational Linguistics (Volume 1: Long Papers)}, 2204--2213.

\end{thebibliography}

\end{document}